\documentclass[letterpaper]{article} 
\usepackage{aaai25}
\usepackage{times}  
\usepackage{helvet}  
\usepackage{courier}  
\usepackage[hyphens]{url}  
\usepackage{graphicx} 
\usepackage{natbib}  
\usepackage{caption} 
\usepackage{algorithm}
\usepackage{algorithmic}

\usepackage{newfloat}
\usepackage{listings}
\DeclareCaptionStyle{ruled}{labelfont=normalfont,labelsep=colon,strut=off} 
\floatstyle{ruled}
\newfloat{listing}{tb}{lst}{}
\floatname{listing}{Listing}
\usepackage[pagebackref=true,breaklinks=true,colorlinks,linkcolor=blue,citecolor=blue,bookmarks=false]{hyperref}

\usepackage{array}
\usepackage{tabularx}
\usepackage{adjustbox}
\usepackage{multirow}
\usepackage{xfrac}
\usepackage{xcolor}
\usepackage{amsmath}
\usepackage{amssymb}
\usepackage{url}
\usepackage{subfigure}
\usepackage{multirow}
\usepackage{booktabs}
\usepackage{threeparttable} 

\usepackage{pifont}
\definecolor{darkergreen}{RGB}{21, 152, 56}
\definecolor{red2}{RGB}{252, 54, 65}
\newcommand{\yesmark}{\textcolor{darkergreen}{\ding{52}}}
\newcommand{\nomark}{\textcolor{red2}{\ding{56}}}

\usepackage{xspace}
\newcommand{\oursfull}{Read, Watch and Scream\xspace}
\newcommand{\ours}{ReWaS\xspace}

\newcommand{\best}[1]{\textbf{#1}}

\usepackage[most]{tcolorbox}

\usepackage{subfigure}
\usepackage{makecell}
\usepackage{xspace}
\usepackage{amsmath}

\newcommand{\bE}{\mathbf{E}}

\def\etal{et al.\xspace}
\def\eg{\emph{e.g}.\xspace}
\def\ie{\emph{i.e}.\xspace}

\newcommand{\myparagraph}[1]{\noindent\textbf{#1}}

\usepackage[capitalize]{cleveref} 
\Crefname{figure}{Figure}{Figure}
\Crefname{table}{Table}{Table}

\title{Read, Watch and Scream! Sound Generation from Text and Video}
\author{
    Yujin Jeong\thanks{Works done during an internship at NAVER AI Lab.}\quad\quad
    Yunji Kim\quad\quad
    Sanghyuk Chun \quad\quad
    Jiyoung Lee\thanks{Corresponding author: lee.j@navercorp.com}
}
\affiliations{
    NAVER AI Lab

}

\begin{document}



\twocolumn[{%
 \renewcommand\twocolumn[1][]{#1}%
\maketitle

\vspace{-12mm}
  \begin{center}
\centering
    \captionsetup{type=figure}
\includegraphics[width=.91\linewidth]{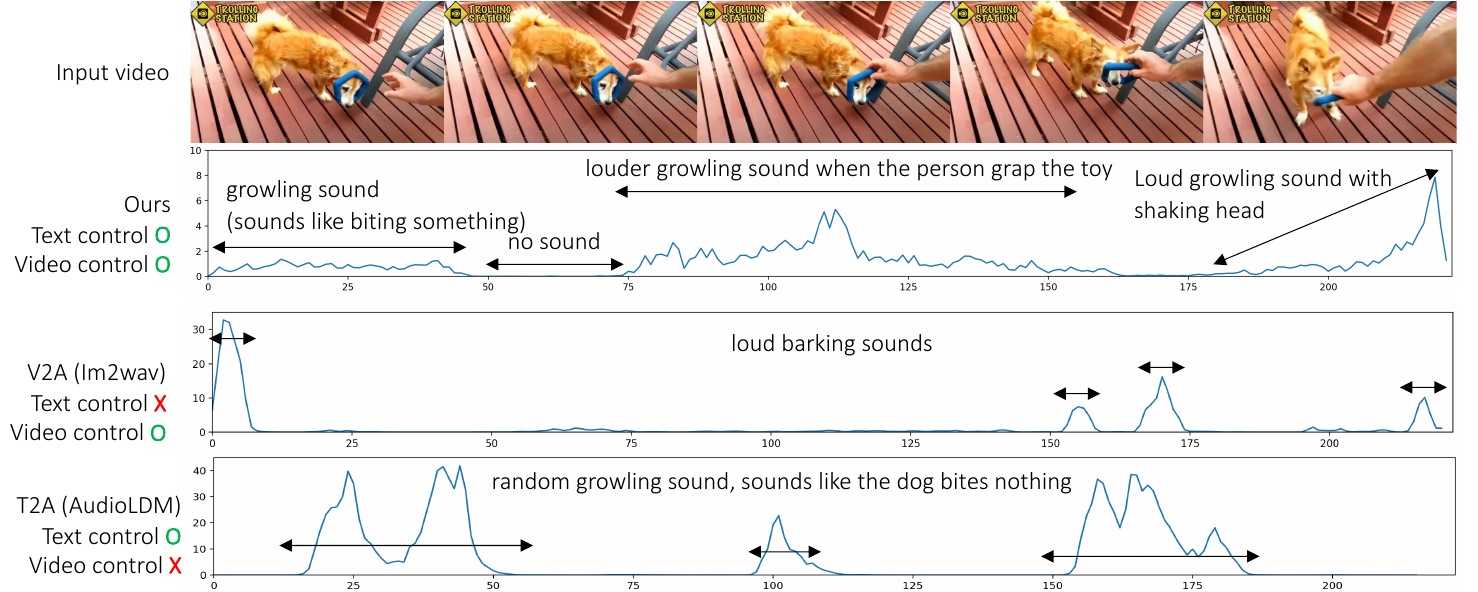} 
  \captionof{figure}{\small An example of audio generation requiring both text and video control. The text instruction ``dog growling'' is used for the text control. The video-to-audio (V2A) \cite{im2wav} or text-to-audio (T2A) \cite{liu2023audioldm} generation methods cannot understand the detailed semantics from texts (the dog is growling, not barking) or video (the dog is biting something, and the alignment), respectively.
}\label{fig:sample-based-comparison-intro}
\end{center}%
 }]

 \renewcommand{\thefootnote}{\fnsymbol{footnote}}
\footnotetext[1]{\footnotesize Work done during an internship at NAVER AI Lab.}
\footnotetext[2]{\footnotesize Corresponding author: lee.j@navercorp.com.}

\begin{abstract}
Despite the impressive progress of multimodal generative models, 
video-to-audio generation still suffers from limited performance and limits the flexibility to prioritize sound synthesis for specific objects within the scene.
Conversely, text-to-audio generation methods generate high-quality audio but pose challenges in ensuring comprehensive scene depiction and time-varying control.
To tackle these challenges, we propose a novel video-and-text-to-audio generation method, called \ours, where video serves as a conditional control for a text-to-audio generation model.
Especially, our method estimates the structural information of sound (namely, energy) from the video while receiving key content cues from a user prompt.
We employ a well-performing text-to-audio model to consolidate the video control, which is much more efficient for training multimodal diffusion models with massive triplet-paired (audio-video-text) data.
In addition, by separating the generative components of audio, it becomes a more flexible system that allows users to freely adjust the energy, surrounding environment, and primary sound source according to their preferences.
Experimental results demonstrate that our method shows superiority in terms of quality, controllability, and training efficiency.
Code and demo are available at \url{https://naver-ai.github.io/rewas} 

\end{abstract}


\section{Introduction}
Generative models have developed dramatically, making content creation easier for people.
Especially, text-to-video generation models such as Make-a-Video~\cite{singer2022make} and Sora~\cite{sora} show the impressive emergence of generative models in the video domain, showing remarkable utility in film and advertising.
While we are fully immersed in the video content by watching and listening, unfortunately, these generated videos are silent.
Generating the sound aligned to a video is a challenging task requiring both a contextual and temporal understanding of the video.
\cref{fig:sample-based-comparison-intro} shows an example of when text and video controls are required to generate a sound that precisely matches the given scene. Here, the dog is growling while holding a toy in his mouth. A human can imagine the sound of the video; the dog growls lowly, and the growling sounds like the dog is biting something. When the person grips and pulls the toy, the dog will treat the human by growling louder. Finally, when the dog shakes his head, the growling will become louder.
If a generative model does not understand the visual information, it will be a random growling sound, not like the dog biting something. If audio is not controlled by text, the generated audio might be only related to the dog, \eg, a barking sound.


\cref{tab:taxonomy_comp} shows the recent attempts to generate an audio sample from the given video or text. There are two major directions to generate an audio sample from the given video directly. First, there have been studies of a sound effect (SFX) generation with short moments for video editing tasks \cite{comunita2024syncfusion,condfoleygen}, known as Foley. They are restricted to the pre-defined sound effect classes and can only control discrete information, such as onset. As another attempt, video-to-audio (V2A) generation methods have been proposed \cite{luo2024diff,xing2024seeing,iashin2021taming,im2wav}. However, they still struggle to generate open-domain sounds from multiple objects together. Furthermore, both SFX and V2A methods cannot take text controls, more rich user control. \cref{fig:sample-based-comparison-intro} shows the example when there is no text control; a V2A method just generates audio of ``barking'' rather than ``growling'' by focusing on the dog in the video. 


As another line of research, text-to-audio (T2A) generation has been actively studied \cite{huang2023make2,makeanaudio1,liu2023audioldm,liu2023audioldm2,ghosal2023tango}. Despite their diverse and high-quality audio generation quality, they lack a temporal understanding of video-only information. Like the example in \cref{fig:sample-based-comparison-intro}, the text-only condition can make irrelevant audio to the video (\eg, when the dog shakes heads). To tackle the problem, we may need more controllability to the T2A model, such as AudioLDM \cite{liu2023audioldm}.
Recently, a few studies \cite{wu2023music, guo2024audio, chung2024t} tried to control the pre-trained AudioLDM more precisely based on ControlNet \cite{controlnet}. Although they can control the pitch, temporal order, energy, or rhythm of the generated audio, their generation process needs expensive timestamp-wise annotations for each control feature.

More recently, parallel to our study, SonicVisionLM~\cite{xie2024sonicvisionlm} and Seeing\&Hearing~\cite{xing2024seeing} incorporate text information, providing users the freedom to generate specific sounds. Although these methods can control audio generation with both vision and language, they still suffer from either limited discrete control (\eg, onset) \cite{xie2024sonicvisionlm}, or lacking timestamp-wise control \cite{xing2024seeing}.
Moreover, they require a video-to-text converting process, such as video captioning or feature mapping, for use with the T2A model. This text conversion weakens temporal alignment, leading to the loss of fine-grained temporal details.

\begin{table*}[ht!]
\centering
\setlength{\tabcolsep}{2pt}
\begin{threeparttable}
\small
\begin{tabular}{cccccc}
\toprule
\multirow{2}{*}{Method} & General Text & Visual & W/o V2T & Efficient \\
& sound? & control? & control? & mapping? & training? \\\midrule
Sound effect generation \cite{comunita2024syncfusion,condfoleygen} & \nomark & \nomark & \yesmark$^\dagger$ & \yesmark & \nomark \\
Video-to-audio \cite{luo2024diff,iashin2021taming,im2wav} & \yesmark$^\ddagger$ & \nomark & \yesmark & \yesmark & \nomark \\
Text-to-audio \cite{huang2023make2,makeanaudio1,liu2023audioldm,liu2023audioldm2,ghosal2023tango} & \yesmark & \yesmark & \nomark & \yesmark & \nomark \\
Text-to-audio \& Control \cite{wu2023music, guo2024audio, chung2024t} & \yesmark & \yesmark & \nomark & \yesmark & \yesmark \\
Video-to-text \& Text-to-audio \cite{xie2024sonicvisionlm,xing2024seeing}  & \yesmark & \yesmark & \yesmark $^\dagger$$^\star$ & \nomark & \yesmark \\
Video-and-text-to-audio (\textbf{\ours}) & \yesmark & \yesmark & \yesmark & \yesmark & \yesmark \\ \bottomrule
\end{tabular}
{
\scriptsize
\begin{tablenotes}
    \item[$\dagger$] Unable to adjust continuous sound variations (\ie, energy). \quad $^\ddagger$ Hardly generate sounds of multiple subjects together.
    \item[$\star$] Taking limited timestamp-wise visual control (\eg, requiring the full timestamp-wise onset annotations, or only able to take a few frames)
\end{tablenotes}
}
\end{threeparttable}
\label{tab:comparison}
\caption{\small Comparison of audio generation methods: Can it make a general sound? Can it take text or visual control? Does it need video-to-text (V2T) mapping? and the training efficiency.}
\label{tab:taxonomy_comp}
\end{table*}

In this work, we propose a novel video-and-text-to-audio generation approach, named \oursfull (\ours), by integrating video as a conditional control for a well-established T2A model.
While a text prompt specifies the subject, we additionally employ a control feature extracted from the video.
More specifically, our method presents an energy adapter on AudioLDM motivated from ControlNet \cite{controlnet}, an efficient structure control method for text-to-image generation. Since a video feature does not directly imply the structure of the audio, we estimate the temporal \textit{energy} information, a basic audio structural information, from the video.

The energy operates as a time-varying control to complement the sound according to the dynamics of the given video.
As shown in \cref{fig:sample-based-comparison-intro}, \ours successfully understands complex information from both text and video.
Here, we define energy as the mean of frequency in each audio frame, which is related to visual dynamics and semantics~\cite{jeong2023power,guo2024audio}.
It is relatively simple to estimate from a video rather than complex acoustic features (\eg, mel-spectrograms). 
Therefore, our energy control facilitates connecting video for T2A model, reflecting strong alignment between audio and video.

We compare our method and other state-of-the-art video-to-audio generation models \cite{condfoleygen,xing2024seeing,luo2024diff,iashin2021taming,im2wav} on two video-audio aligned datasets, VGGSound \cite{chen2020vggsound} and GreatestHits \cite{greatesthits}. In the experiments, 
\ours outperforms V2A methods in human evaluation for three categories (audio quality, relevance to the video, and temporal alignment between audio and video) with a significant gap (almost +1 point for every category in 5-scale MOS).
Also, \ours shows a superior audio generation performance quantitatively and qualitatively. Our method shows the best fidelity score (FD), structure prediction (energy MAE), and AV-alignment score on VGGSound. Moreover, we achieve the best AP and energy MAE on Greatest Hits without the use of reference audio samples like CondFoleyGen \cite{condfoleygen}. As shown in the qualitative study, 
\ours can capture the challenging ``short transition'' of the video when the skateboarder jumps into the air, and no skateboarding sound appears in the video.
It is also possible to generate the sound of a video generated by a general text prompt.

\section{Related Work}
\subsection{Text-to-audio generation}
Early work for audio generation was built upon GANs \cite{kreuk2022audiogen,dong2018musegan}, normalizing flows \cite{kim2020glow}, and VAEs \cite{van2017neural}. 
Recently, several studies using diffusion models have shown promising progress on a broad range of acoustic domains. 
DiffSound~\cite{yang2023diffsound} employs a diffusion-based token decoder for the first time to transfer text features into mel-spectrogram tokens. 
Make-An-Audio~\cite{makeanaudio1}, AudioLDM~\cite{liu2023audioldm}, AudioLDM2~\cite{liu2023audioldm2}, Tango~\cite{ghosal2023tango} and Make-An-Audio2~\cite{huang2023make2} are well-founded in latent diffusion model (LDM)~\cite{ldm}, demonstrating high-quality results with large scale training. 
A series of LDM predicts mel-spectrograms using a VQ-VAE decoder, and a pretrained vocoder generates raw waveforms from the generated mel-spectrograms.
While these methods successfully generate high-quality audio samples for the given text prompt, they are only designed for taking text conditions, unable to understand visual semantics.

Meanwhile, there have been a few attempts based on ControlNet \cite{controlnet},
an efficient training method for structure control for text-to-image generation. 
ControlNet utilizes hints (\eg, Canny edge maps, scribbles, depth maps) to provide a structural composition to the generated images. 
Inspired by this, text-to-audio methods have incorporated ControlNet to accomplish controllable  music~\cite{wu2023music} and audio effect~\cite{guo2024audio,chung2024t} generation.
They have provided more explicit and fine-grained control over the generated audio, leading to performance improvement and adherence to the desired characteristics. 

However, designing these time-varying controls still requires costly labor for users.
To address this challenge, we predict energy control through a given video, which is a practical function for creating SFX, post-production for filmmaking, and utilizing AI-generated silent videos.

\subsection{Video-to-audio generation}
Existing video-to-audio (V2A) generation methods have focused on achieving two main characteristics: (i) audiovisual relevance and (ii) temporal synchronization. 
The first stream aims to represent general sound by leveraging datasets such as VGGSound~\cite{chen2020vggsound} and AudioSet~\cite{audioset}. 
Given a set of video features, SpecVQGAN~\cite{iashin2021taming} learns a transformer to sample quantized representations (\ie, codebook) based on visual features to decode spectrogram. 
Im2wav~\cite{im2wav} utilizes rich semantic representations obtained from a pre-trained CLIP~\cite{clip} as sequential visual conditioning for an audio language model, and applies CFG \cite{ho2022classifier} to steer the generation process. 
Recently, diffusion-based models have shown the stunning ability to generate high-quality audio~\cite{luo2024diff,xing2024seeing}.
DiffFoley~\cite{luo2024diff} improves audiovisual relevance by learning temporal and semantic alignment through contrastive learning.
However, it necessitates tremendous training data, such as the utilization of both VGGSound and AudioSet for alignment training.
Seeing\&hearing~\cite{xing2024seeing} is another diffusion-based model that optimizes the text-to-audio diffusion model, AudioLDM~\cite{liu2023audioldm} by using ImageBind~\cite{girdhar2023imagebind} which learns joint embedding space for six modalities (image, text, audio, depth, thermal, and IMU).
However, ImageBind Video Encoder takes only two frames for each video sampled from 2 second, which results in lacking timestamp-wise contol.
Therefore, they often struggle to generate temporally aligned sounds at short times in the video (\eg, dog barking, people laughing).

On the other hand, other research works~\cite{comunita2024syncfusion,xie2024sonicvisionlm} have focused on creating simplistic SFX (\eg, stick hits) using datasets like CountixAV \cite{countixav} and GreatestHits \cite{greatesthits}, which provide fewer classes but more precisely temporal aligned data. 
CondFoleyGen~\cite{condfoleygen} trains a Transformer to autoregressively predict a sequence of audio codes for a spectrogram VQGAN, conditioned on the given audiovisual example. 
Syncfusion~\cite{comunita2024syncfusion} predicts a discrete onset label that denotes the beginning of a sound for repetitive actions.
Recent SonicVisionLM~\cite{xie2024sonicvisionlm} employs a large language model to utilize text as an intermediate product that facilitates user interaction for personalized sound generation. 
They freeze Tango~\cite{ghosal2023tango} and train ControlNet with timestamp estimated by a video for 23 SFX categories exclusively, where the video is converted to sound event timestamp and text.
Although they have shown promising results in SFX generation, their timestamp detection module is limited to a single visual object, and they cannot implicit detailed temporal cues in visual content because they use videos to convert them into text.
our method generates sounds for various categories from the visual context at the same time.

\section{Preliminary}
\subsection{Text-to-audio latent diffusion model}\label{subsec:audioldm}
In this paper, we specifically utilize AudioLDM~\cite{liu2023audioldm} which generates a latent of mel-spectrogram $z$ computed by VAE \cite{kingma2013vae}.
The diffusion model $\epsilon_\theta$ of AudioLDM is trained to predict the noise added to a given data by minimizing the objective function, $\mathcal{L}_{\text{diff}}=\mathbb{E}_{z_0, \epsilon, t} \left\Vert \epsilon - \epsilon_\theta(z_t, t, \mathbf{E}_{a}) \right\Vert^2_2$,
where $\epsilon$ represents the noise added at time $t$, $z_t$ is noisy latent induced via the forward process and $\mathbf{E}_{a}$ denotes the embedding of the audio $x$ obtained from the CLAP audio encoder $f_\text{audio}(\cdot)$ \cite{laionclap2023}. Here, the model is conditioned by $\mathbf{E}_{a}$ using classifier free guidance (CFG) \cite{ho2022classifier}.

In the sampling process, the generation starts from a noise $z_T$ sampled from $\mathcal{N}(0, I)$ and the text embedding $\mathbf{E}_y$ from the CLAP text encoder $f_\text{text}(\cdot)$. 
The reverse process conditioned on $\mathbf{E}_y$ generates the audio prior $z_0$ using the modified noise estimation $\hat{\epsilon}_\theta(z_t, t, \mathbf{E}_y) =  (1 + w)\epsilon_\theta(z_t, t, \mathbf{E}_y)-w\epsilon_\theta(z_t, t)$,
where $w $ is a guidance weight to balance the audio condition $\mathbf{E}_{a}$. 
The VAE decoder decodes the sampled latent $z$ to predict a mel-spectrogram. Finally, the decoded mel-spectrogram is converted to a raw audio sample using the HiFi-GAN vocoder \cite{hifigan}.

Although AudioLDM enables text-conditional audio generation, it still lacks of understanding of visual contents and their temporal information. This study adds a visual control to the pre-trained AudioLDM.
Instead of directly using a visual feature to control, we extract more essential information from the given video, which will be discussed in \cref{sec:temp}.

\subsection{Video-to-audio with temporal alignment}\label{sec:temp}

\begin{figure}
    \centering
    \begin{minipage}{.52\linewidth}
    \includegraphics[width=\linewidth]{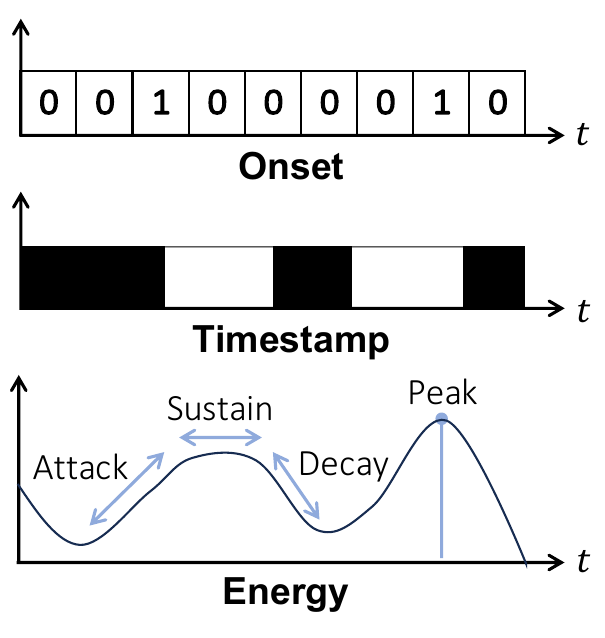}
    \caption{\small Discrete timestamp annotations vs. Continuous energy.}
    \label{fig:energy}        
    \end{minipage}\hfill
    \begin{minipage}{.44\linewidth}
    \includegraphics[width=\linewidth]{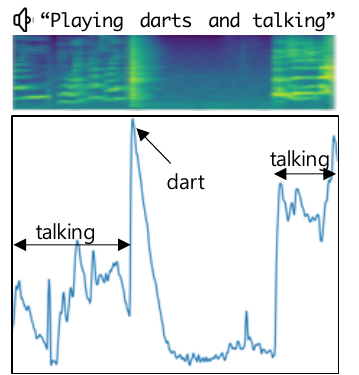}
    \caption{\small Energy can improve temporal alignment.}
    \label{fig:energy_example}
    \end{minipage}
\end{figure}

We assert that a video input can bring principal temporal information that is hard to convey with a text prompt. 
However, directly injecting temporal information from visual into an audio generation model remains a significant challenge.
In contrast,
previous works have attempted to generate sound by estimating the onset~\cite{comunita2024syncfusion}, or audio timestamp~\cite{xie2024sonicvisionlm} from videos to improve audiovisual relevance.
However, they are limited to producing an unnatural sound for a single object in that discrete conditions cannot serve continuous sound variations.

In this work, we consider \textit{energy}, the averaged mel-spectrogram on the frequency axis, to produce a continuous condition.
\cref{fig:energy} shows that energy is a continuous time-varying signal, including envelope components of sound such as peak, attack, sustain, and decay.
Energy can be obtained cheaply and automatically by computing the frame-level magnitude of mel-spectrograms~\cite{ren2020fastspeech}.
Moreover, we empirically observe that energy can also implicitly improve the temporal alignment of the video. For example, \cref{fig:energy_example} shows energy can contain continuously varying audio information.

\section{Method}
This paper introduces a novel sound generation method conditioned on text and video, to generate a waveform temporally well aligned with the visual input. 
Our model consists of two parts: 
(i) {\em control prediction}, which intermediately predicts energy control from the video. (\cref{sec:control}) 
(ii) {\em conditional sound generation}, which uses the energy control signal as a condition in the diffusion process to generate corresponding audio outputs (\cref{sec:generation}), which are both temporally and semantically aligned with text and video.

\begin{figure*}[t]
\begin{center}
    \includegraphics[width=.85\linewidth]{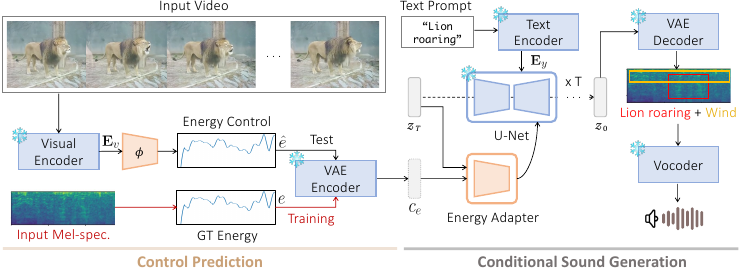}
    \caption{\small Overall architecture of \ours. Our model predicts energy control from a given video, and generates sound with text prompt and control condition. Red lines are used in training only, and replaced to the video-to-energy estimator $\phi$ in test time.}
    \label{fig:overview}
\end{center}
\end{figure*}

\subsection{Energy control prediction from video}\label{sec:control}

\myparagraph{Energy control.}
\ours is based on AudioLDM that uses CLAP embeddings for text and audio alignment. A na\"ive approach using video as a condition is to align latent space between audio-video-text. 
\citet{luo2024diff} attempted to align tri-modal embeddings in a unified space by large-scale contrastive learning prior to training diffusion models.
To more efficiently overcome this challenge, we design an energy control as an intermediate bridge from video to audio.
We speculate that energy control brings three advantages: First, the power of audio is intuitively correlated to visual dynamics and semantics~\cite{jeong2023power,sung2023sound}. With the natural fact that people can imagine the power of sound from the size of the instance or distance to the object, we regard audio energy as a visually correlated signal that can be certainly obtained from video.
Second, as shown in \citet{ren2020fastspeech} and \citet{guo2024audiocontrol}, energy plays as a structural condition for audio generation. Thus, it is well-suited to parameter-efficient fine-tuning methods such as ControlNet. 
Finally, using temporal acoustic signals for generating audio needs a skilled user to annotate the pitch, melody, or rhythm for every timestamp. It makes the audio generation phase impractical and difficult for the public to control. Meanwhile, energy is highly related to physical interactions implicated in visual signals; thus, it can be easily estimated from the video. Our approach does not require timestamp-wise fine-grained user control, but automatically estimating energy structure from the given video.

\noindent\myparagraph{Video embedding.}
To predict the energy control from video input, we extract features from the pretrained SynchFormer \cite{iashin2024synchformer} video encoder.
We empirically observe that the image encoder (\eg, CLIP~\cite{clip}) is limited to V2A generation, especially from a temporal alignment perspective.
We finally take video embedding $\bE_v\in \mathbb{R}^{S \times C}$, where $S$ is the number of segments and $C$ is the dimension of latent.
The implementation details for this process are described in Appendix.

\myparagraph{Training energy control from video.}
Similar to Ren \etal\cite{ren2020fastspeech}, we calculate the energy from the mel-spectrogram by averaging the frequency bins and further smoothing the time-sequential energy information. 
We first transform the raw waveform to the mel-spectrogram, $\texttt{mel} \in \mathbb{R}^{D \times W}$, where $D$ represents the number of mel-frequency bins, and $W$ is the width of the spectrogram following AudioLDM~\cite{liu2023audioldm}.
However, we empirically observe that the computed energy fluctuates a lot for each temporal window, which hinders stable training.
We resolve the issue by taking a smoothing operator. The energy of audio $e \in \mathbb R^W$ is defined as $e_a = \texttt{Smoothing} \left(\frac{1}{D} \sum_{d=1}^{D} \texttt{mel}_{w,d}\right)$.
We use the second-order Savitzky-Golay filter~\cite{virtanen2020scipy} with a window length of 9 for smoothing.

We estimate $\hat{e}$ by using a shallow projection module $\phi$ from the video encoder output (See \cref{fig:overview} ``Control Prediction''). For efficient training, we resize $e_a$ by taking the nearest-neighbor interpolation to have the same number of segments $S$ as the visual representations. We also can apply the same resize method to video embeddings at inference time. Now, we train our energy control prediction module $\phi$ by minimizing the following loss function 
$\mathcal{L}_{e} =||\phi(\bE_v) - \texttt{Resize}({e})||_2^2$.

The output $\hat{e}$ of the projection module is used for energy control at inference time. 
We train $\phi$ separately to diffusion models for training efficiency.
In addition, our energy estimation module is not specialized for generation models, thus our energy control can be utilized in other ways.

\subsection{Conditional sound generation}\label{sec:generation}
\myparagraph{Adding control signal.}
To reflect the energy control signal, we train the energy adapter following ControlNet~\cite{controlnet}.
The weights of the energy adapter are initialized from pretrained parameters of diffusion models, and connected to AudioLDM with zero convolution layers.
Compared to training audiovisual alignment into the latent space in diffusion model~\cite{luo2024diff,xing2024seeing}, our adapter takes the benefit of robust fine-tuning speed (\eg, \citet{luo2024diff} uses 8 A100 GPUs for 140 hours for feature alignment and LDM tuning, whereas we use 4 V100 GPUs for total 33 hours).
To add the control feature for $z_t$, the energy control $e_a$ is duplicated by the number of mel-filterbanks, and transferred to the VAE encoder for the purpose of encoding, followed by a fully-connected layer and SiLU \cite{silu}.
This latent control feature $c_e$ is added to the $z_0$, where $z_0$ is an audio prior obtained from the VAE encoder.
Thus, given a text embedding $\bE_y$ and latent control feature $c_e$, we train energy adapter by optimizing the following objective:
$\mathcal{L}_{c} = \mathbb{E}_{z_0, t,  \mathbf{E}_y, c_e, \epsilon \sim \mathcal{N}(0, 1)} \|\epsilon - \epsilon_\theta(z_t, t, \mathbf{E}_y, c_e)\|_2^2.$
During training, we randomly drop $\mathbf{E}_y$ with the probability 0.3 for better controls.
We denote that $\mathcal{L}_c$ and $\mathcal{L}_e$ are optimized separately.

\myparagraph{Sound generation.} 
We use DDIM \cite{ddim} to generate sound from the noise. 
The reverse sampling process is conditioned on both text and video.
We replace $e$ to $\hat{e}=\phi(\bE_v)$ at inference.
Once mel-spectrogram is generated by the VAE decoder, it can be transformed into a raw waveform using the pre-trained vocoder~\cite{hifigan} (See \cref{subsec:audioldm}).
Note that conditioning on video increases the total inference time by only 3\%.

\section{Experiments}
\subsection{Experimental settings}
\myparagraph{Datasets.}
For a fair comparison with existing baselines, we train the control prediction module and the adapter in the conditional sound generation module on VGGSound~\cite{chen2020vggsound}. 
VGGSound is a large-scale dataset containing $\approx$200k video clips, accompanied by corresponding audio tracks. 
The dataset covers 309 classes of general sounds, roughly categorizing them into acoustic events, music, and people. 
The videos are sourced from YouTube, providing a diverse and realistic corpus. 
Since the VGGSound includes plentiful general sound examples, \ours trained on the VGGSound enables general-purpose sound generation for real-world scenarios.
We randomly sampled 3K videos to construct VGGSound test subset.
To evaluate temporal alignment accuracy, we use Greatest Hits~\cite{greatesthits} test set including the videos of hitting a drumstick with materials. Since Greatest Hits samples have a distinct audio property compared to the other audio samples, we fine-tune \ours on the Greatest Hits training samples.


\myparagraph{Baselines.}
We compare \ours against open-source V2A generation approaches in priority, SpecVQGAN~\cite{iashin2021taming}, Im2wav~\cite{im2wav} and Diff-Foley~\cite{luo2024diff}, which are trained on the VGGSound and AudioSet datasets. 
Furthermore, we compare Seeing\&Hearing~\cite{xing2024seeing}, which optimizes a pre-trained AudioLDM during the inference stages by aligning the latent space using ImageBind. 
For a fair comparison, we take the following steps: 
We first generate the full-length audio by each method, and use a common 5-second clip for evaluation.
In the temporal alignment evaluation,
we consider CondFoleyGen~\cite{condfoleygen} as a main baseline, which is trained on the Greatest Hits dataset.

\myparagraph{Evaluation metrics.}
Following the implementation of AudioLDM, we employ Fréchet distance (FD) \cite{fid}, Fréchet audio distance (FAD)~\cite{fad}, and the mean of KL divergence (MKL)~\cite{iashin2021taming}.
We also measure the alignment between the generated audio and sound categories with CLAP score~\cite{makeanaudio1}.
However above metrics are limited to evaluating audio-visual temporal alignment, so we employ AV-align~\cite{yariv2024diverse} based on detecting energy peaks in audio-visual modalities.
In the Greatest Hits experiment, we report onset accuracy (Acc) and average precision (AP), following the evaluation protocol introduced by CondFoleyGen.
The onset of sound events is a discrete signal obtained by the thresholding of the amplitude gradient.
Therefore, relatively quiet sound effects (\eg, scratching leather, touching the leaves) or natural sounds can be excluded from the evaluation.
To address this issue, we report the mean absolute error (MAE)~\cite{guo2024audio} of the energy signals from real and generated sounds for the first time in the sound generation task conditioned on video. 
Although these evaluation metrics can evaluate different properties of the generated audio, most of them measure the difference between the generated audio and the ``ground truth'' audio corresponding to the original video. However, one video can sound differently (\eg, human's voice can vary); existing quantitative evaluation metrics have challenges in measuring whether the generated audio is truly suited to the given video. To tackle the issue, we conduct a user study to evaluate the quality and temporal alignment of the generated audio samples. 

\subsection{Results}

\begin{table}[t]
    \footnotesize
    \centering
        \resizebox{\linewidth}{!}{
        \setlength{\tabcolsep}{2.5pt}
        \begin{tabular}{l cccc ccc}
        \toprule
        {Model}              
        &{FD$\downarrow$} 
        & FAD$\downarrow$
        & MKL$\downarrow$
        & {CLAP$\uparrow$} 
        & {MAE$\downarrow$} 
        & {AV-align$\uparrow$} 
        & {\# TP$\downarrow$} \\ 
        
        \midrule
        SpecVQGAN
        &     26.63    &    5.57  &   3.30   
        &    0.1336     &    0.1422  & 0.2851 &   379M \\ 
        
        Im2wav
        &   16.87     &     5.94   &   2.53
        &   0.4001     &     0.1310   & 0.2763 &   365M   \\
           
        Diff-Foley
        &   21.96  &  6.46  &  3.15    
        &   0.4010   &     0.1571     &  0.2059 & 859M  \\ 

        Seeing\&Hearing
        &   20.72   &   6.58   &   \best{2.34}      
        &   \best{0.5805}  &    0.1668   & 0.1858     &  -  \\ 
        
        \midrule

        \ours (Ours) 
         & \best{15.24}    & \best{2.16}    &  2.78     
        &  0.4353     & \best{0.1149}  & \best{0.3008} & \best{204M} 
        \\
           
        \bottomrule
        
        \end{tabular}%
        }
        \caption{\small Performance comparison on VGGSound~\cite{chen2020vggsound} with reproduced five seconds audio samples. ``Energy'' and ``TP'' denote energy MAE and number of the trainable parameters.
    }
    \label{tab:vggsound}
\end{table}

    
        
        
           

        

           
        

\begin{table}
    \small
    \centering
    \resizebox{.7\linewidth}{!}{
    
    \begin{tabular}{l ccc}
            \toprule
            
            Model  
            & Acc$\uparrow$ 
            & AP$\uparrow$    & MAE$\downarrow$ \\
            
            \midrule
            
            CondFoleyGen
            & \best{23.94} & 60.24 & 0.1520\\ 

            \ours (Ours)
            & 19.15 & \best{63.28} & \best{0.1398} \\ 
            
            \bottomrule 
        \end{tabular}
        }
        \caption{\small Performance comparison on Greatest Hits~\cite{greatesthits}. We use material types as text prompts, while CondFoleyGen uses both reference audio and video as inputs.}
    \label{tab:comp_greatesthits}
\end{table}

\myparagraph{Quantitative results.}
\cref{tab:vggsound} shows the quantitative comparisons on the VGGSound.
We note that category classes are used as text prompts in the VGGSound.
We train 22M parameters for video projection to audio conditional control, and 182M parameters for fine-tuning the AudioLDM with our energy adapter. 
Since Seeing\&Hearing is an optimization-based generation method, we did not report the training parameters. However, they consume twice the time for inference than \ours.
Our \ours achieve the best performance on FD, FAD, energy MAE, and AV-align, showing competitive results in terms of MKL and CLAP scores.
Especially, while we use only a quarter of training parameters compared to Diff-Foley, our method outperforms Diff-Foley on all metrics.  
CLAP scores illustrate the importance of text prompts for semantic alignment.
Seeing\&Hearing outperforms \ours in terms of MKL and CLAP score. However, we argue that Seeing\&Hearing is heavily dependent on text prompt, since our method outperforms in terms of MAE and AV-align scores by a large margin.
This achieved MAE score result by \ours also demonstrates the accuracy of our control prediction module, and generated outputs from \ours are most temporally closer to the real audio content.

In addition, we evaluate how the generated audio and the condition video are temporally aligned on Greatest Hits. The dataset distribution of Greatest Hits highly differs from the general audio samples; hence, we fine-tune \ours on the Greatest Hits training samples. \cref{tab:comp_greatesthits} shows the results. \ours achieves the best AP and MAE, although \ours is not specially designed for Foley like CondFoleyGen. 



\begin{table}
\small

\centering
\resizebox{\linewidth}{!}{
\setlength{\tabcolsep}{2pt}
\begin{tabular}{lccc}\toprule
Model  & \makecell{Audio Quality $\uparrow$} & Relevance $\uparrow$  & \makecell{Temporal Alignment  $\uparrow$}\\\midrule
SpecVQGAN & 2.76 & 2.50 & 2.64 \\
Im2wav & 2.97 & 3.18 &  3.01 \\
Diff-Foley & 2.89 & 2.97 & 2.98 \\ 

\midrule

\ours (Ours) & \textbf{3.70} & \textbf{4.04} & \textbf{3.68} \\
\bottomrule
\end{tabular}
}
\caption{\small Human evaluation of V2A methods on audio quality, audiovisual relevance, and temporal alignment with 5-scale MOS.}\label{tab:human_eval}
\end{table}
\myparagraph{User study.}
The quantitative results are limited to measuring how the generated audio sounds realistic and aligned to the given video.
To complement it, we conduct a human evaluation study to assess the subjective quality of the generated audio concerning the input video.
We ask the human evaluators to evaluate the quality of the audio samples generated by SpecVQGAN, Im2wav, Diff-Foley, and \ours.
Since Seeing\&Hearing shows vulnerable performance in audio-visual alignment, we exclude it from the user study.

We evaluate three criteria: audio quality, relevance between audio and video, and temporal alignment. 
We use a five-point Likert scale to measure mean opinion score (MOS), where an ideal video with its ideal audio receives a rating of 5 across all criteria.
We recruit human annotators via two separate channels: Amazon Mechanical Turk (AMT) and local hiring. We recruit 50 AMT annotators for each criterion, and each annotator evaluates five generated samples for each method (\ie, each annotator evaluates 20 audios). Locally hired 23 annotators evaluated 20 generated samples for each method and criterion. 
Surprisingly, \ours achieves the best in all categories with large margins as shown in \cref{tab:human_eval}.
This subjective result is consistent with our quantitative and qualitative findings, further validating the effectiveness of \ours in generating high-quality, relevant, and temporally synchronized audio for the given video.

\begin{figure}
\centering
      \includegraphics[width=.95\linewidth]{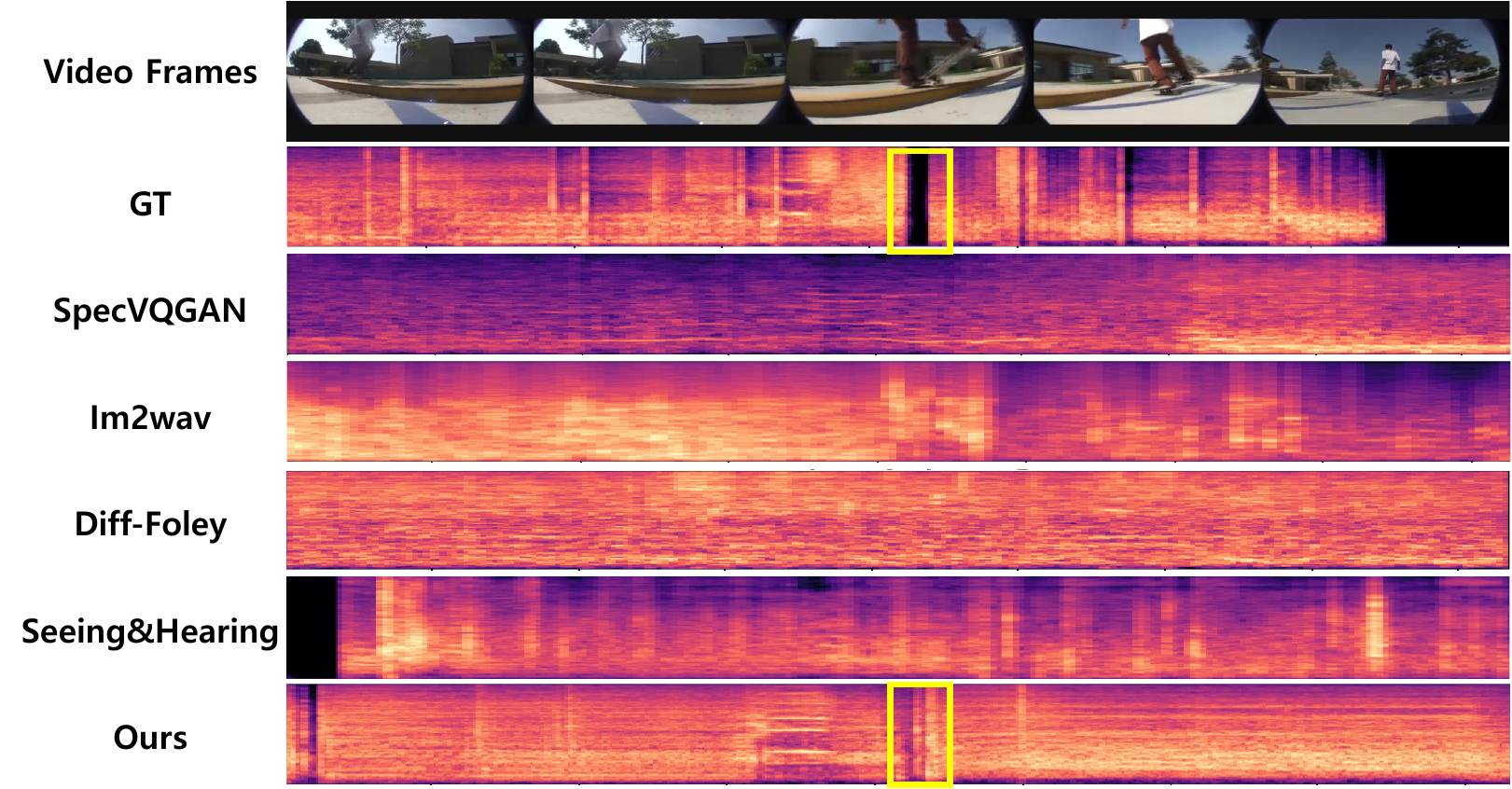}

  \caption{\small 
  Qualitative comparison on VGGSound.
  Surprisingly, when the skateboarder jumps, only \ours succeded in detecting short transition (yellow box).  Text prompt in is ``skateboarding''.}
  \label{fig:qualitative}

\end{figure}

\begin{figure}[t]
\begin{center}
  \includegraphics[width=\linewidth]{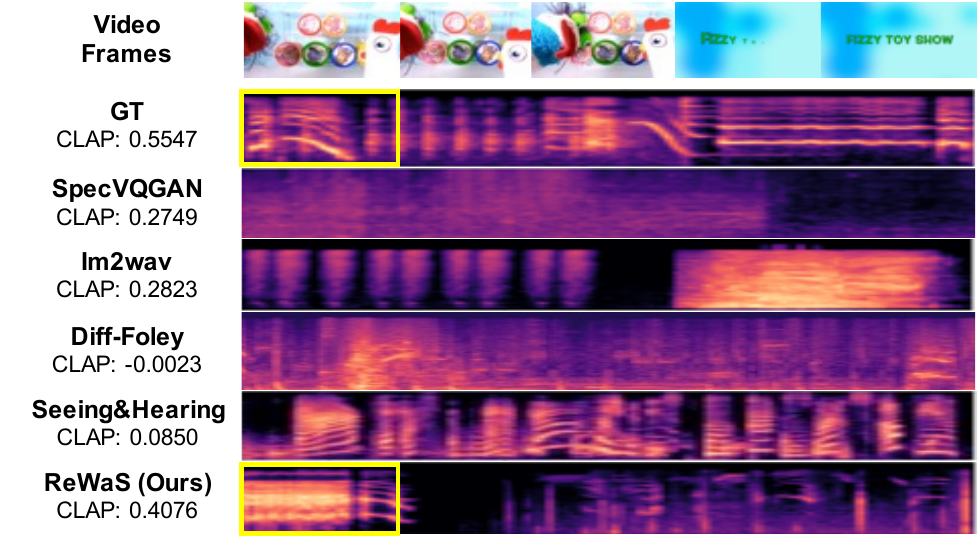}
  \caption{\small Effectiveness of text prompt. When videos contain both semantic and redundant frames, text prompts used in \ours calibrate the results. Text prompt in is ``chicken clucking''}
  \label{fig:texteffect} 
\end{center}
\end{figure}

\myparagraph{Qualitative results.}
\cref{fig:qualitative} shows qualitative results in baselines and \ours. Given the skateboarding video, SpecVQGAN and Diff-Foley fail to generate the sound of skate wheels rolling on the floor. Although Im2wav generates that sound, it cannot capture a short transition.
We also demonstrate the effectiveness of the text prompt in \cref{fig:texteffect} with CLAP similarity, when redundant frames exist.
In this case, V2A methods also struggle to generate corresponding sound.
However, \ours can effectively calibrate the semantics by user text prompt. Note that the prompt can also be longer and more general if desired by users. (\eg, ``rally car swiftly navigates a turn on the racetrack.")
\begin{table}[t]
    \centering
    \small
    \resizebox{\linewidth}{!}{
    \begin{tabular}{lcccccc}
        \toprule
       Control & Backbone & FD$\downarrow$ & FAD$\downarrow$ & KL$\downarrow$ & CLAP$\uparrow$ & MAE $\downarrow$ \\
        \midrule
        T \& GT E from A
        & AudioLDM-M & 13.93 & 2.65 & 2.15 & 0.4497 & 0.1195\\
        T \& Est. E from V
        & AudioLDM-M & 15.24 & 2.16 & 2.78 & 0.4353 & 0.1149 \\
        T \& Est. E from V
        & Make-An-Audio & 13.89 & 10.91 & 2.93 & 0.4237 & 0.1368 \\
        \bottomrule
        \end{tabular} 
        }
        \caption{\small Impact of the energy control's quality on VGGSound. (1) Text and the ground-truth audio energy with AudioLDM backbone (upper bound), (2) Text and the estimated energy from the video with AudioLDM (our approach) and (3) with Make-An-Audio.
    }\label{tab:condition}
\end{table}

    

\subsection{Discussion}

\myparagraph{The impact of the quality of the energy control.}
To verify the robustness of the energy prediction module, we compare the control by our video-to-energy prediction module and the energy directly extracted from the ground truth audio. \cref{tab:condition} demonstrates that although we use the estimated energy, the quality of the generated audio is very similar to the audio samples controlled by the ground truth audio energy. (See the first row and second row of \cref{tab:condition})
It supports the idea that energy information is highly related to visual information, and is easy to estimate solely using video.
We also compare the qualitative results of estimated energy controls with ground truth energy in the \cref{sec:app_qual}.

\myparagraph{T2A framework.}
We replace the AudioLDM-M backbone with Make-An-Audio~\cite{makeanaudio1}, which has fewer parameters than AudioLDM to validate the flexibility of our approach. 
Table~\ref{tab:condition} shows the results of the two backbones on VGGSound.
Interestingly, \ours built upon Make-An-Audio achieves comparable performance to its AudioLDM-M counterpart, demonstrating the robustness of our framework (See the second row and third row of \cref{tab:condition}).




\begin{figure}[t]
\begin{center}
  \includegraphics[width=1\linewidth]{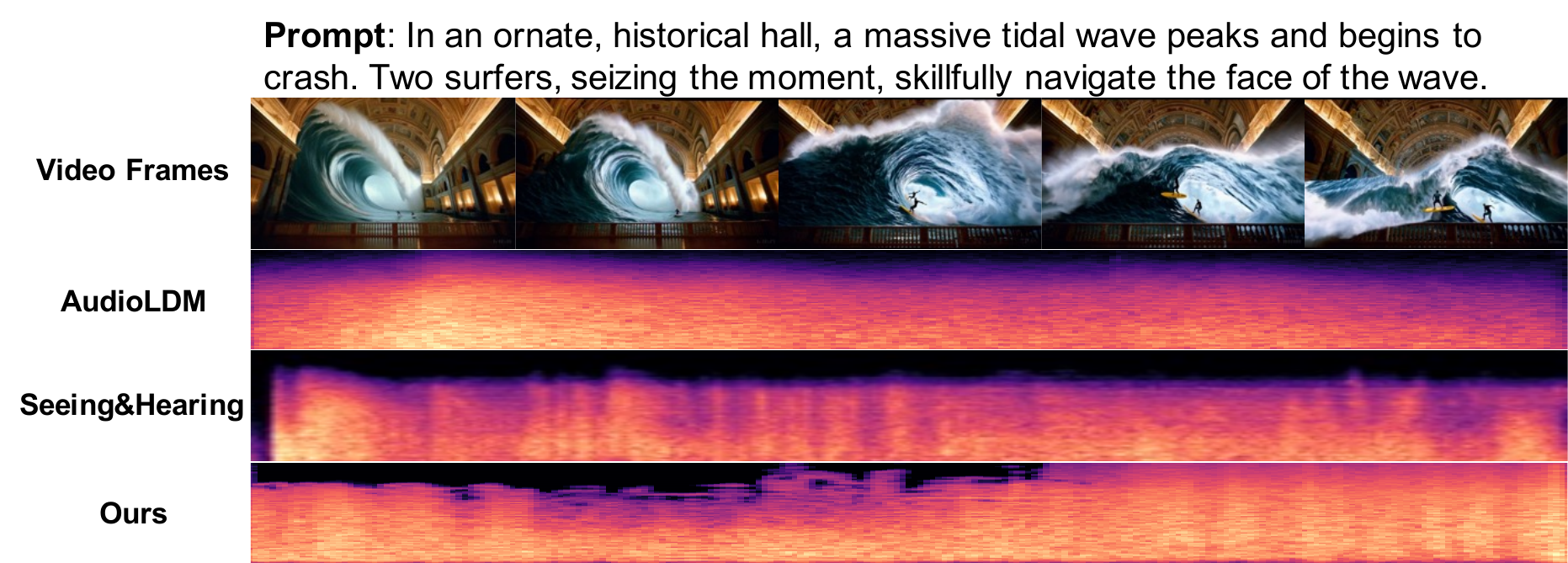}
   \caption{\small Audio generation with general text prompts.}
  \label{fig:general_prompt} 
\end{center}
\end{figure}

\myparagraph{General text prompts.}
To examine the capability of \ours with more general text prompts, we generate audio samples with generative videos by KLING\footnote{https://kling.kuaishou.com/en}.
As shown in \cref{fig:general_prompt}, only our method can capture the visual information of the wave, namely, the sound getting louder as the wave crashes.
We include more samples in the \cref{sec:app_qual} and demo.

\myparagraph{Effectiveness of visual condition.} 
In the Appendix, we show examples when energy control complementing temporal information. While AudioLDM suffers from inferior temporal alignment and limited sound generation that is mentioned in text prompts but not generated, \ours not only generates video-related sounds hidden in the text but also aligns the sound with the frames. This demonstrates the effectiveness of visual condition by \ours. 
More detailed discussion is in the \cref{sec:app_qual}.



\section{Conclusion}
This paper proposes \ours, a novel video-and-text-to-audio generation framework. Our key idea lies in inferring audio structural condition, namely energy,  from video to efficiently and effectively input the visual condition to the robust T2A model.
Therefore, \ours can generate complex sounds in the real world without the need for a difficult control design.
Quantitative results on VGGSound and Greatest Hits datasets, subjective human study, and qualitative results consistently support that \ours can generate natural, temporally well-aligned, and relevant audio for the given video by employing text and video as control.

\section*{Acknowledgements}
The NAVER Smart Machine Learning (NSML) platform~\cite{kim2018nsml} had been used for experiments.

\appendix
\section{Appendix}\label{sec:app}
\subsection{Data preprocessing}\label{app:data}

During training, we randomly extract 5-second segments from the VGGSound dataset~\cite{chen2020vggsound} and 2-second segments from the Greatest Hits dataset. 
However, during the testing phase, we extract video clips ranging from 2 to 7 seconds in duration for the VGGSound dataset, and from 0 to 2 seconds for the Greatest Hits dataset~\cite{greatesthits}. 
Video frames are uniformly sampled at 25 fps.
Since \ours generates audio based on 5-second videos, we duplicated frames from the Greatest Hits dataset to match the length of these 5-second videos. 
Subsequently, we trimmed the generated audio to a duration of 2 seconds.

For comparison with baselines, SpecVQGAN~\cite{iashin2021taming}, Diff-Foley~\cite{luo2024diff} and Seeing\&Hearing~\cite{xing2024seeing} (10s, 8s, and 10s, respectively) for the test videos.
Then, we extract the 5-second clip corresponding to the same video frames used in our method. 
Since Im2wav~\cite{im2wav} is designed to generate sound with a fixed length of 4 seconds, we first generate the initial 4 seconds and extend it by generating an additional 1 second, resulting in a 5-second audio clip.

\subsection{Feature extraction}\label{app:feat}
\myparagraph{Video features.}
We employ SynchFormer~\cite{iashin2024synchformer} trained on VGGSound~
\cite{chen2020vggsound} for the sparse synchronized setting as a video encoder.
The video encoder employed in SynchFormer is based on Motionformer~\cite{patrick2021keeping} pre-trained on Something-Something v2~\cite{goyal2017something}, and fine-tuned on VGGSound and AudioSet~\cite{audioset}.
Therefore, the video encoder is strong enough to encode motion dynamics and semantics.
We freeze the parameters in the video encoder, and solely train a projection module to estimate energy control.
We extract a video feature in the short video clip (0.64 sec).
Thus we use a total of 112 length visual embeddings for a 5s video.
We note that, for a fair comparison, RGB frames are only used in all methods including \ours. 

\myparagraph{Audio features.}
Audios of all videos used in our experiments are resampled to 16kHz sampling rate.
We follow the default setting of AudioLDM to compute the mel-spectrogram.
Specifically, we use 64-bin mel-spectrograms with 1024 window length. While $f_{\text{min}}$ and $f_{\text{max}}$ are 0 and 8000 respectively, the hop size is 160 and the FFT size is 1024.


\subsection{Architecture and training details}\label{app:details}
\myparagraph{Test dataset.}
For our experiments, we leverage a subset of 160k videos from VGGSound~\cite{chen2020vggsound} due to the availability of public videos at the time of training. We split the train data list into training and validation subsets following SpecVQGAN~\cite{iashin2021taming}. 

\myparagraph{Energy signal.}
To encode a video feature into 1-dimensional energy, a projection module $\phi$ consists of a linear layer, two transformer blocks, and MLPs consisting of four FC layers.
We use 768 hidden dimensions for the first linear layer and transformer blocks, and the four FC layers' output dimensions are 128, 64, 16, and 1.
The total parameter of $\phi$ is 22M.
We choose AudioLDM-M$\footnote{weights in \url{https://github.com/haoheliu/AudioLDM}}$, and the number of training parameters for fine-tuning AudioLDM~\cite{liu2023audioldm} with our energy adapter is 182M.
\ours is optimized by AdamW and the learning rate is fixed to 3e-5 during training.
We train \ours with 4 V100 GPUs for 33 hours on VGGSound, and 1 hour on Greatest Hits~\cite{greatesthits} respectively.


\myparagraph{Details of Make-An-Audio backbone framework.}
The video encoder used in Make-An-Audio~\cite{makeanaudio1} is re-trained to predict the appropriate energy scale of mel-spectogram, which is configured with 80 frequency bins and a hop size of 256 samples, different from the AudioLDM-M configuration. Make-An-Audio is notable for its parameter efficiency, requiring significantly fewer parameters than AudioLDM. This reduction in model complexity translates to substantially shorter training times, with the entire model converging in less than one day.

\subsection{User study}
\label{subsec:appendix_user_study}
\cref{fig:inst1}, \cref{fig:inst2}, and \cref{fig:inst3} show the user instructions used in our human evaluation.
Before launching Amazon MTurk (AMT), we first conducted an in-lab study with 23 participants; each participant evaluated 20 audio samples for each method and each criterion, namely, they evaluated 240 (20 $\times$ 4 $\times$ 3) generated audio samples.
Based on the observation from the in-lab study, we have set the compensation level for each HIT to \$0.45 so that a worker can earn \$15 per hour.
At the same time, we observed that a number of participants had trouble keeping focus on the evaluation with 240 samples (each sample takes five seconds). To prevent the low-quality responses from MTurk annotators, we split each evaluation Human Intelligence Task (HIT) on a smaller scale.
Each AMT annotator evaluates five audio samples for each method and one additional ground truth audio to prevent random guessing.
We published 50 HITs for each criterion, and 150 responses were collected.
Finally, we observe that many AMT annotators consistently score high for all questions (\eg, 4 or 5). To ignore noisy responses, we omit responses having an average score larger than 4.0 for 21 questions. 55 responses were omitted after this filtering process.

\newtcbtheorem{Instruct}{\bfseries Instruction}{enhanced,drop shadow={black!50!white},
  coltitle=black,
  top=0.3in,
  attach boxed title to top left=
  {xshift=1.5em,yshift=-\tcboxedtitleheight/2},
  boxed title style={size=small,colback=pink}
}{instruction}

\newtcolorbox[auto counter]{instruction}[1][]{title={\bfseries Instruction~\thetcbcounter},enhanced,drop shadow={black!50!white},
  coltitle=black,
  top=0.3in,
  attach boxed title to top left=
  {xshift=1.5em,yshift=-\tcboxedtitleheight/2},
  boxed title style={size=small,colback=pink},#1}

\begin{figure}[h]
    \centering
    \begin{Instruct}{}{first}
        How natural is this audio recording?
         \\ \\
        Please focus on examining the audio quality and naturalness (noise, timbre, sound clarity, and high-frequency details).
        \\
        \\
        1. Listen to the sample  (Click **Play** button to listen audio samples) \\
        2. Select an option
        \begin{itemize}
            \item Excellent: 5 (Completely natural audio)
            \item Good: 4 (Mostly natural audio)
            \item Fair: 3 (Equally natural and unnatural audio)
            \item Poor: 2 (Mostly unnatural audio)
            \item Bad: 1 (Completely unnatural audio)
        \end{itemize}
        \vspace{0.5em}
    \end{Instruct}
\caption{User instruction for audio quality (naturalness) test.}\label{fig:inst1}

\end{figure}

\begin{figure}
    \centering
    \begin{Instruct}{}{second}
        How much is the sound related to the object or material in video?
        \\ \\
        Please focus on examining the relevance between video and audio, not considering the quality and temporal alignment (\ie sound timing).
        \\
        \\
        1. Watch the sample (Click **Play** button to watch video samples) \\
        2. Select an option 
        \begin{itemize}
            \item Excellent: 5 (Completely relevant audio)
            \item Good: 4 (Mostly relevant audio)
            \item Fair: 3 (Equally relevant and irrelevant audio)
            \item Poor: 2 (Mostly irrelevant audio)
            \item Bad: 1 (Completely irrelevant audio)
        \end{itemize}
        \vspace{0.5em}
\end{Instruct}
\caption{User instruction for video-audio relevance test}\label{fig:inst2}
\end{figure}

\begin{figure}
    \centering
    \begin{Instruct}{}{third}
        How much is the sound temporally aligned to the movements of objects or material in video?\\ \\
        Please focus on examining the temporal alignment between video and audio, not considering audio quality and naturalness.
        \\
        \\
        1. Watch the sample (Click **Play** button to watch video samples) \\
        2. Select an option
        \begin{itemize}
            \item Excellent: 5 (Completely aligned audio)
            \item Good: 4  (Mostly aligned audio)
            \item Fair: 3  (Equally aligned and non-aligned audio)
            \item Poor: 2  (Mostly non-aligned audio)
            \item Bad: 1  (Completely non-aligned audio)
        \end{itemize}
        \vspace{0.5em}
    \end{Instruct}
\caption{User instruction for temporal alignment test.}\label{fig:inst3}
\end{figure}

\subsection{More qualitative results}\label{sec:app_qual}
\begin{figure*}[t]
\begin{center}
      \includegraphics[width=.8\linewidth]{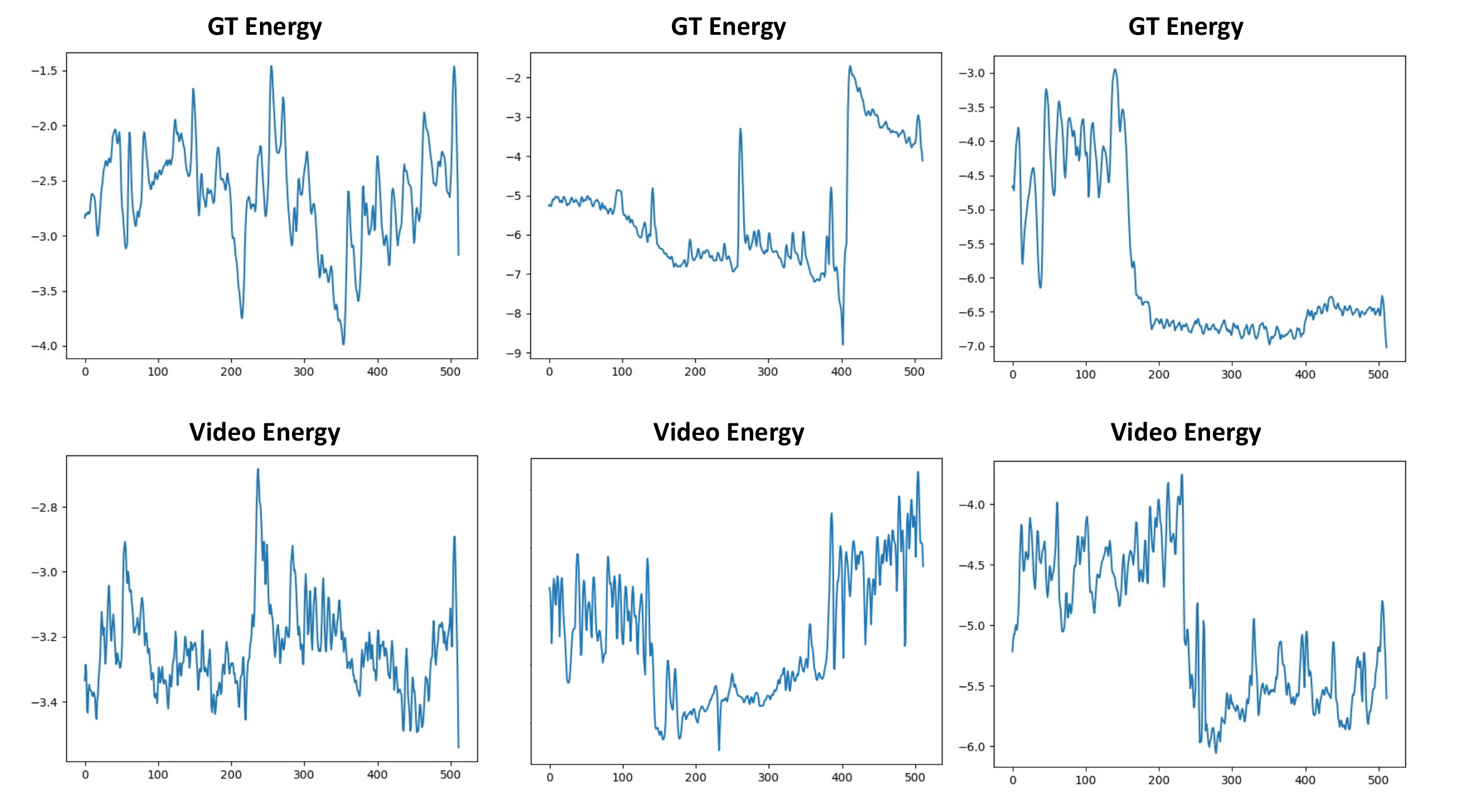}
  \caption{Examples of energy controls from input videos.}
  \label{fig:energy_examples} 
\end{center}
\end{figure*}

\myparagraph{Energy controls from videos.}
We illustrate estimated energy from video in \cref{fig:energy_examples}. The results show the correlation between our energy control generated from video and GT energy obtained from reference audio. 

\begin{figure*}[t]
\begin{center}
  \includegraphics[width=.7\linewidth]{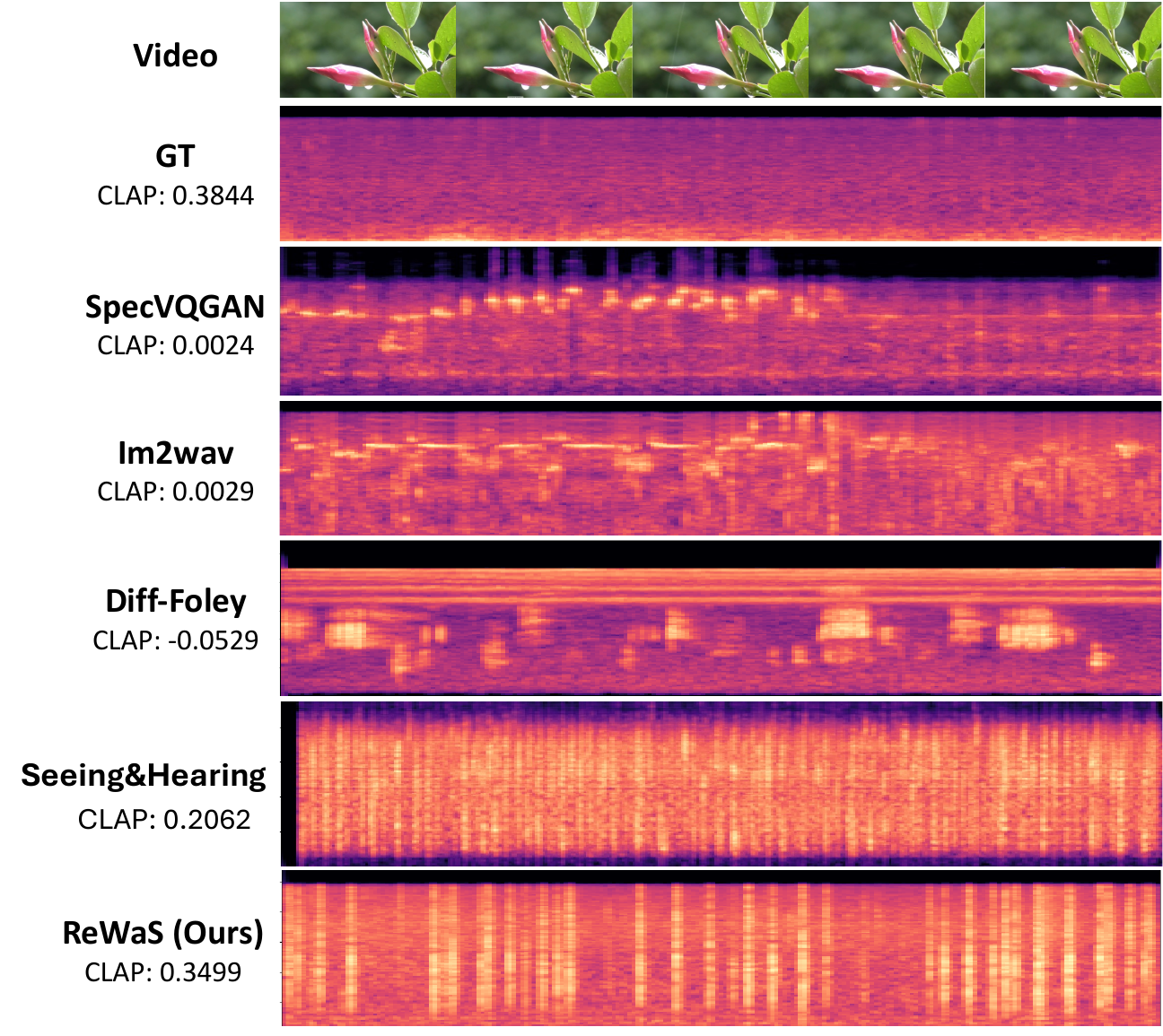}\vspace{-.6em}
  \caption{\small Effectiveness of text prompt. Videos in the real world are sometimes noisy. For example, when videos are hard to distinguish the semantics, text prompts used in \ours calibrate the results. Text prompt is ``raining''.
  }
  \label{fig:difficult_a} 
\end{center}
\vspace{-1em}
\end{figure*}

\begin{figure*}[t]
\begin{center}
    \includegraphics[width=0.8\linewidth]{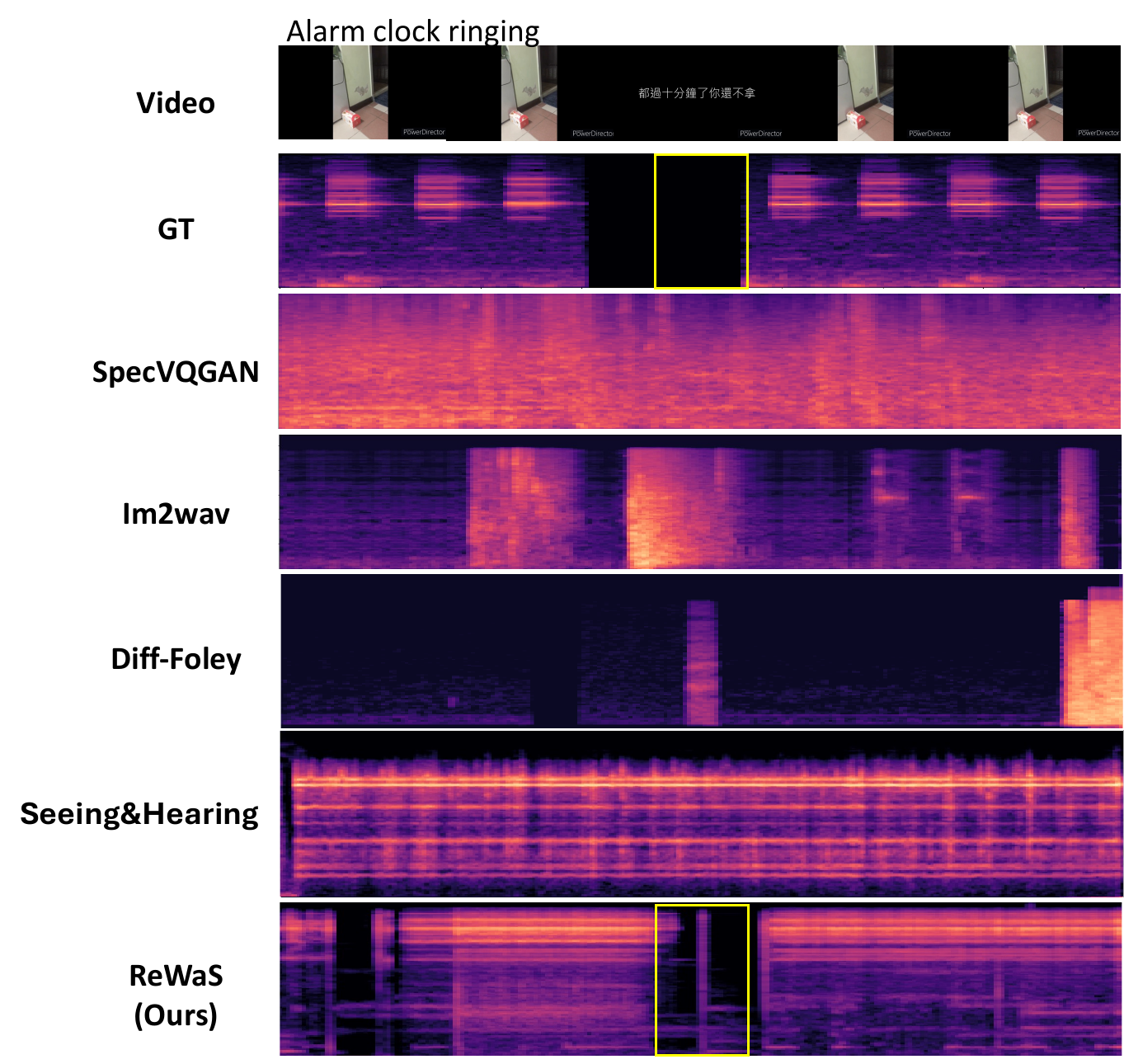} \\
  \caption{Example of audio sound from misaligned visual input. \ours can make the desired sound and make the silent moment like ground-truth sound.}
  \label{fig:silent} 
\end{center}
\end{figure*}

\myparagraph{Effectiveness of the text prompt.}
In the input video shown in Fig. \ref{fig:difficult_a}, rain streaks are barely visible, while we want to hear the sound of rain. \ours can emphasize the desired sound with the help of a text prompt `\textit{raining}'. While V2A methods struggle to generate corresponding sound, Seeing\&Hearing~\cite{xing2024seeing} and \ours can effectively calibrate the semantics by user text prompt. However, \ours shows better CLAP score~\cite{laionclap2023} than Seeing\&Hearing. Furthermore, we assert that Seeing\&Hearing is heavily dependent only on text prompt. For example, as shown in Figure~\ref{fig:silent}, if there are redundant frames, ReWaS can only successfully calibrate the semantics with textual prompt but also it generates ``silent" audio sounds when there is a scene change. In contrast, other baseline models such as SpecVQGAN~\cite{iashin2021taming}, Im2wav~\cite{im2wav} and Diff-Foley~\cite{luo2024diff} fail to produce the corresponding sound (e.g., alarm clock ringing) due to misaligned visual and sound contexts, often generating unintended sounds or remaining silent when they should produce sound. Although Seeing\&Hearing~\cite{xing2024seeing} can produce corresponding sounds, it fails to generate ``silent" audio when there is a change in visual scenes. 
This suggests that baseline models may either resort to generating random sounds when faced with a scene change due to misaligned visual and sound contexts, or they produce sound when they should remain silent, ignoring the visual context.

\begin{figure*}[t]
\begin{center}
    \includegraphics[width=1\linewidth]{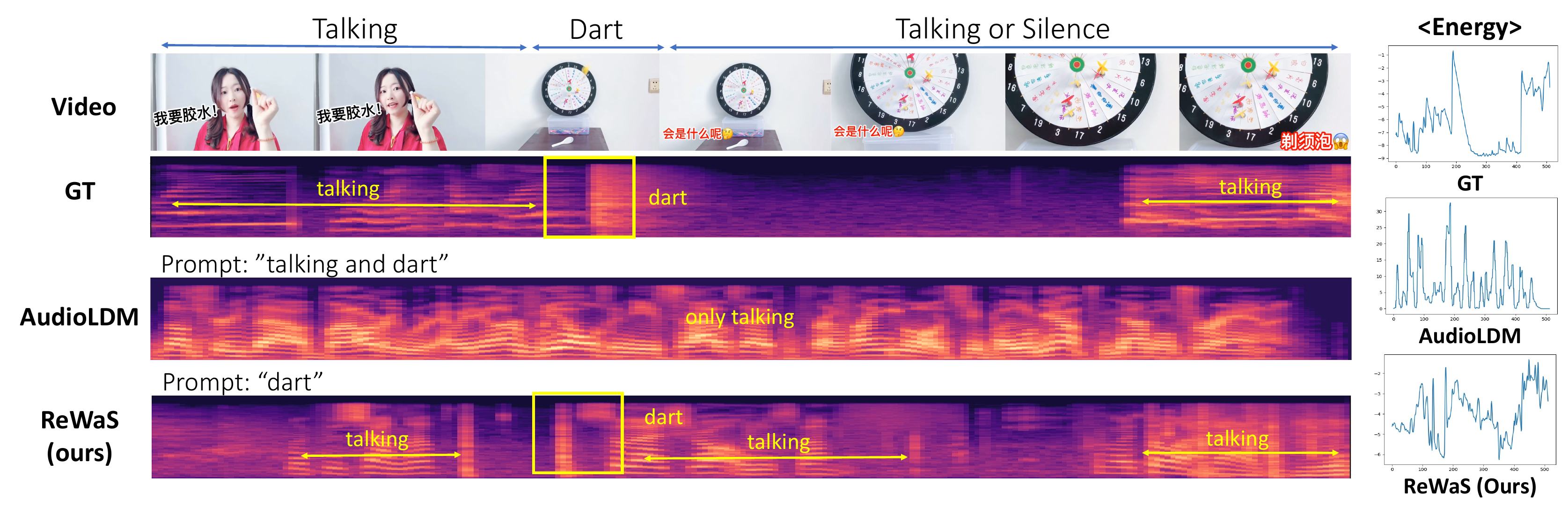} \\
  \caption{Effectiveness of video input. In \ours, energy control from video input transfers additional temporal information.}
  \label{fig:multi_supple} 
\end{center}
\end{figure*}

\begin{figure*}[t]
\begin{center}
    \includegraphics[width=1\linewidth]{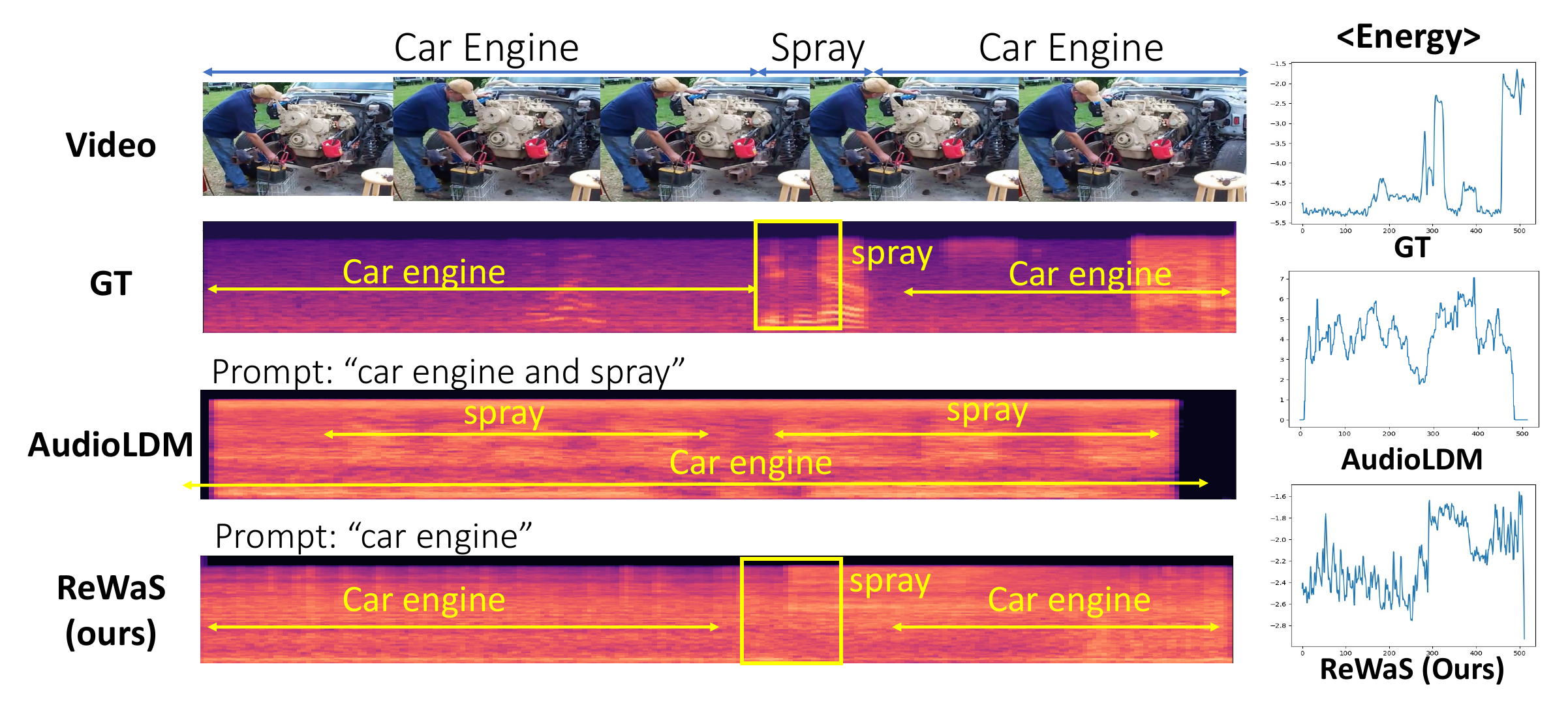} \\
  \caption{Additional example of effectiveness of video input.}
  \label{fig:multi_supple2} 
\end{center}
\end{figure*}

\myparagraph{Effectiveness of visual control.}
\cref{fig:multi_supple} and \cref{fig:multi_supple2} are examples when energy signal serving additional temporal information.
As shown in ~\cref{fig:multi_supple}, when a person talking and playing a dart game in an input video, the original AudioLDM~\cite{liu2023audioldm} generates only the sound of talking, ignoring `dart' prompt. 
Additionally, aligning generated sound with video is challenging in AudioLDM.
In comparison, \ours not only generates both the sound of talking and dart but also aligns the sound with the frames. 
Figure~\ref{fig:multi_supple2} presents another example. Unlike AudioLDM, which repeatedly generates the same spray and car engine sounds, \ours accurately captures the spray sound at the right moment thanks to the visual control without additional text prompt `spray'.
Furthermore, the result demonstrates the limitation of T2A methods for automatic Foley synthesis, because they cannot watch a video. 
This demonstrates the effectiveness of visual control by ReWaS.

\begin{figure*}[t]
\begin{center}
    \includegraphics[width=0.8\linewidth]{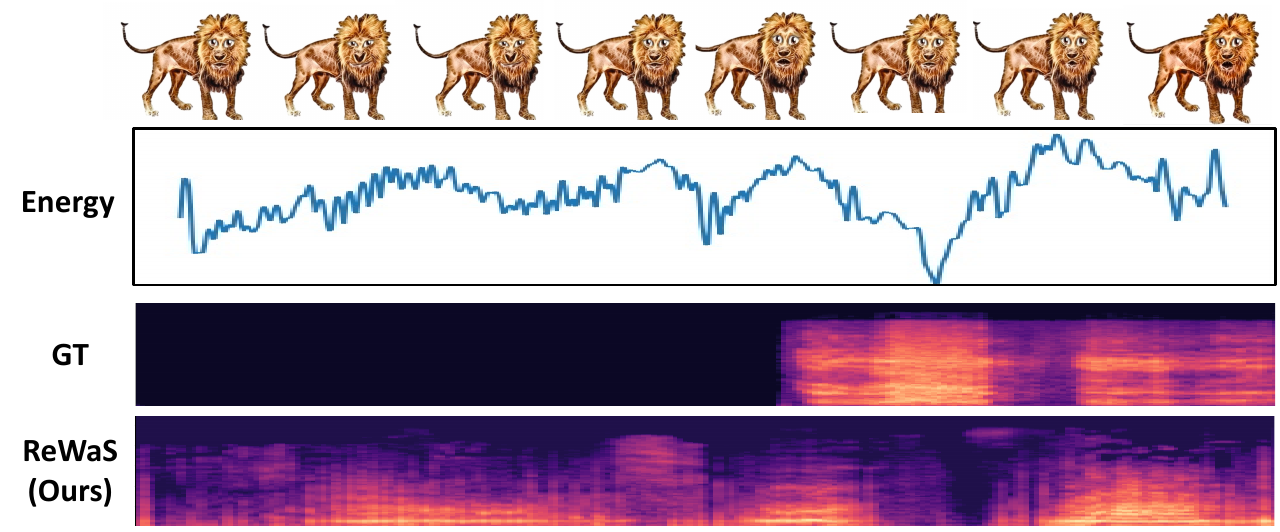} \\
  \caption{Example of audio with improved synchronization, capturing small movements (e.g., a lion's lip synchronization).}
  \label{fig:misaligned} 
\end{center}
\end{figure*}
\begin{figure*}[t]
\begin{center}
    \includegraphics[width=0.7\linewidth]{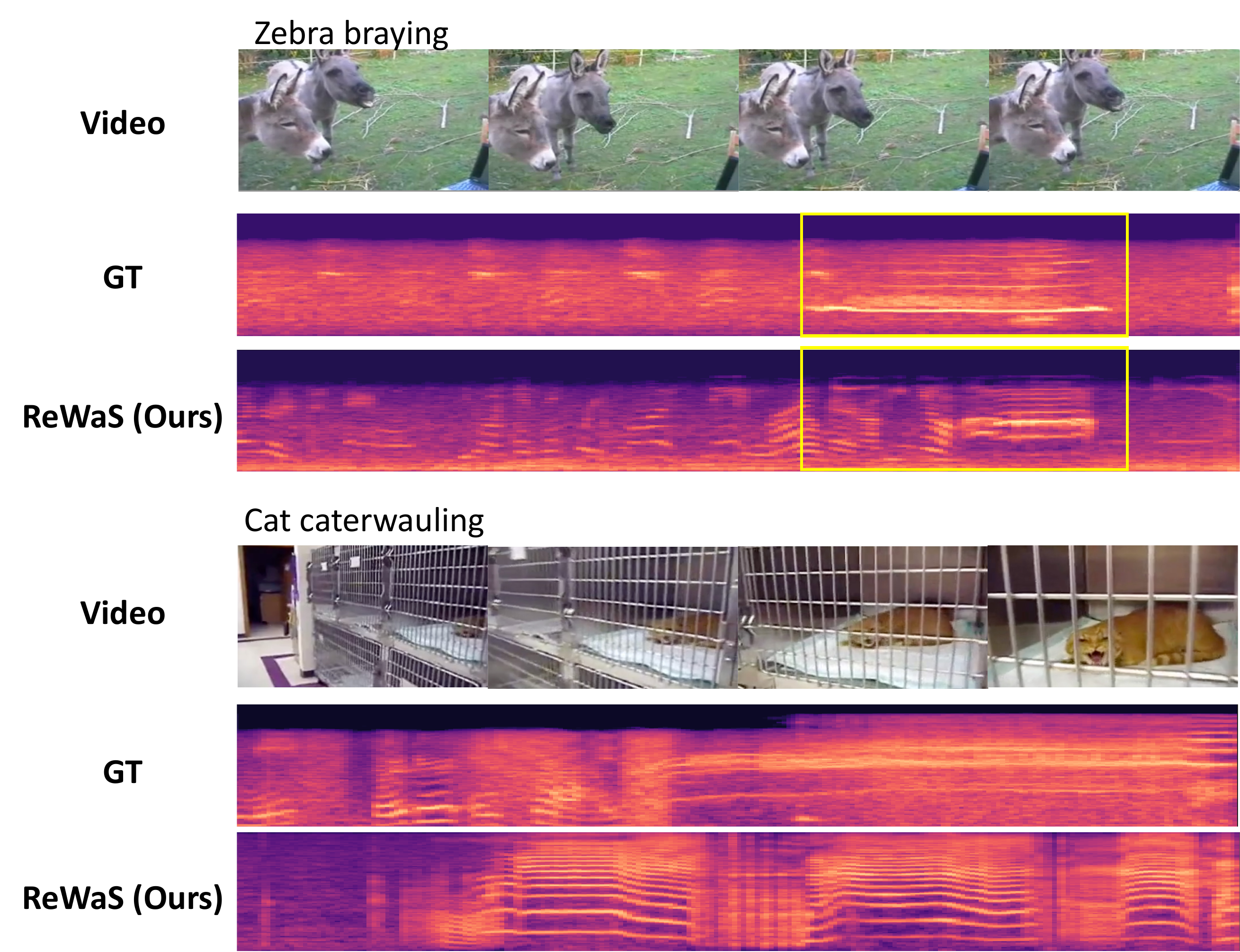} \\
  \caption{Examples of generated audio sounds demonstrating the capability of temporal synchronization.}
  \label{fig:tempo2} 
\end{center}
\end{figure*}

\myparagraph{Subtle visual movements.}
Figure~\ref{fig:misaligned} and Figure~\ref{fig:tempo2} demonstrate the effectiveness of \ours in aligning sound with corresponding frames, achieving temporal alignment by accurately capturing small object movements, such as lip synchronization. As illustrated in Figure~\ref{fig:misaligned}, the the intensity of growling sound increases as the lion opens its mouth. In another example of Figure~\ref{fig:tempo2}, \ours also produces temporally synchronized sounds with mouth movements, underscoring its overall effectiveness.

\begin{figure*}[t]
\begin{center}
    \includegraphics[width=0.7\linewidth]{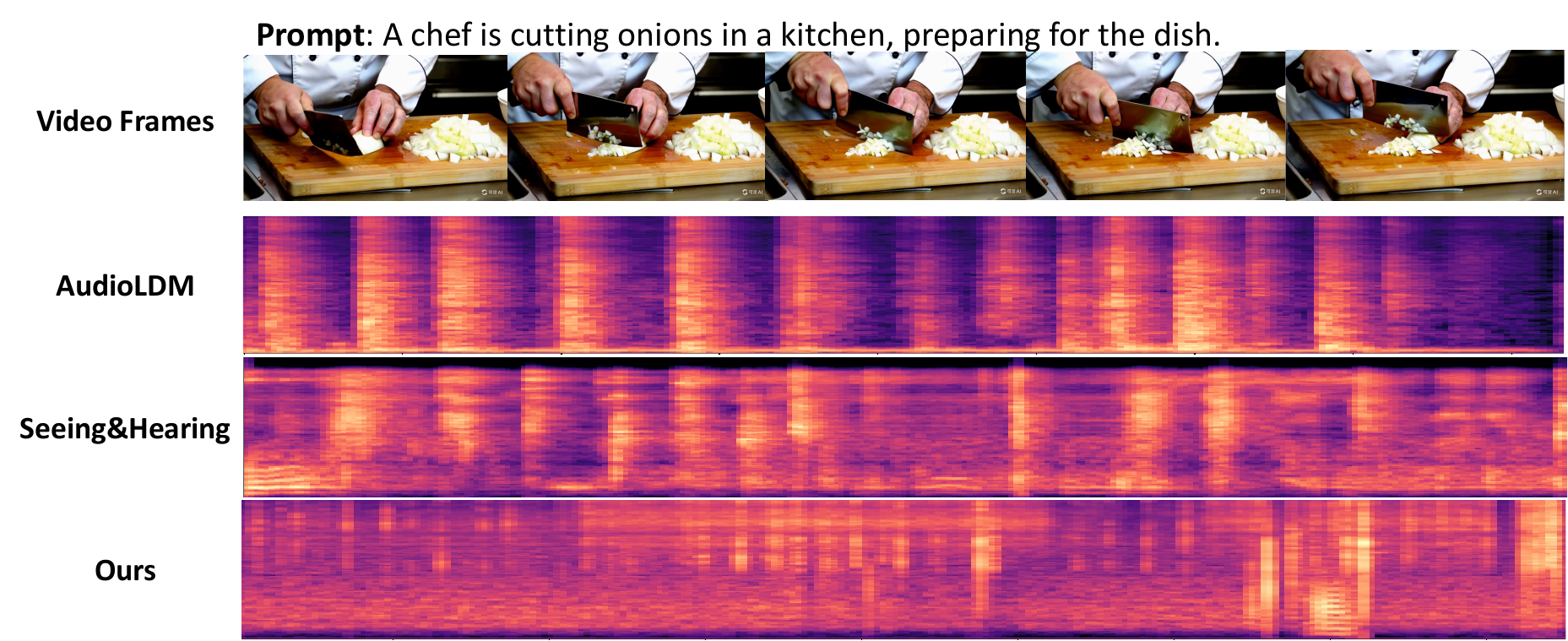} \\
  \caption{Example of general user prompt.}
  \label{fig:general_supple} 
\end{center}
\end{figure*}

\myparagraph{General text prompt.}
Figure~\ref{fig:general_supple} provides an example that evaluates the capability of \ours using more general text prompts. 
We generate audio samples with another generated video from KLING\footnote{https://kling.kuaishou.com/en}.
Our method is the only one that captures the increasing intensity of the sound as the onions are cut from the edge to the center. Both the T2A model, AudioLDM, and the V2T\&V2A model, Seeing\&Hearing, can generate corresponding sounds, but they lack visual temporal alignment in the generated results.

\subsection{Extended analysis}
\begin{figure*}[t]
\begin{center}
    \includegraphics[width=0.7\linewidth]{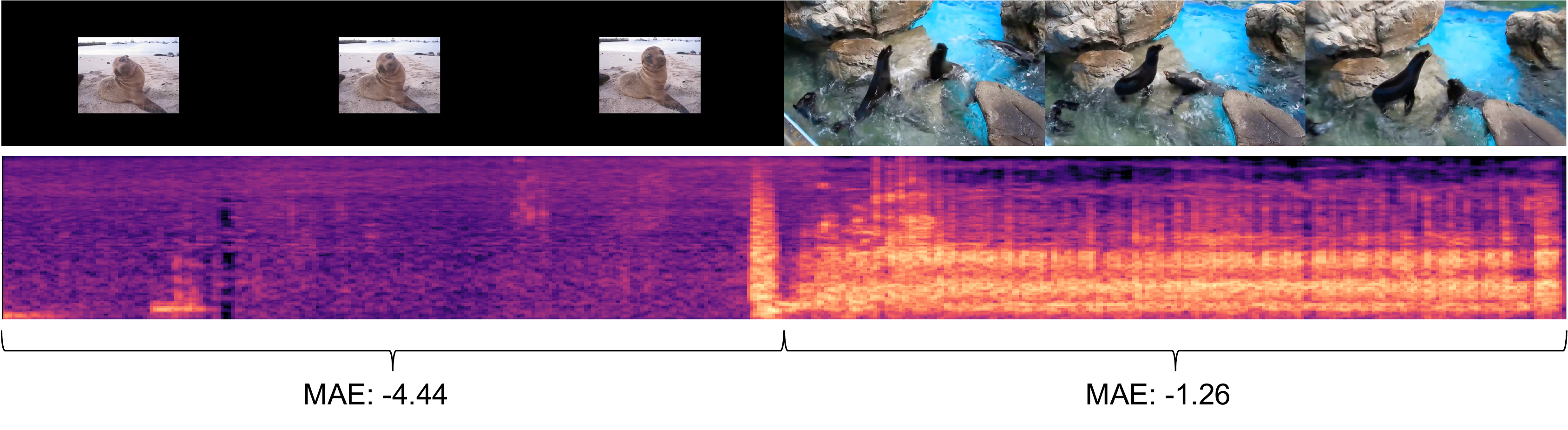} \\
  \caption{Example of concatenation of different video event.}
  \label{fig:different_video} 
\end{center}
\end{figure*}

\myparagraph{Isolating and editing energy across different events.}
We observe that visual controls provide temporal information, raising the following question of whether it is possible to isolate the energy associated with each event or edit the sound on a single track. To address this, we combine two videos with distinctly different styles within the same category. As shown in \cref{fig:different_video}, we observe that the generated sounds change their rhythm or tempo accordingly and average energy MAE across events shows 2-3 times difference, indicating that isolating or editing energy by event is feasible.

\myparagraph{Roles of text prompts and visual energy control.}
In \cref{tab:additional_ablation}, we present additional experimental results on subset of VGGSound~\cite{chen2020vggsound} to explore the specific roles of text prompts and visual energy control. The first and second columns show results using null text and misaligned text, respectively, while the third column shows results without visual energy control.
For misaligned text prompts, we randomly shuffle text prompts from different video categories.
We observe that the model typically generates the main audio sound based on the text prompt while aligning the sound's temporal dynamics with the video movement. 
In particular, misaligned or null text prompts result in a much lower CLAP score~\cite{makeanaudio1}, while the AV-align~\cite{yariv2024diverse} score shows only a slight decrease from 0.30 to 0.27, highlighting the importance of accurate text guidance.
In contrast, omitting energy control results in significant drops in the AV-align score, demonstrating that energy control ensures temporal alignment between generated audio and video.
In conclusion, the text and energy controls each play critical and distinct roles in generating semantically and temporally aligned audio outputs.

\begin{table}[t]
    \small
    \caption{\small Ablation Studies on text prompt and energy controls.}
    \label{tab:additional_ablation}

    \centering
    \resizebox{\linewidth}{!}{
    
    \begin{tabular}{l cccc}
            \toprule
            
            Metrics & Ours & w/o Text& misaligned Text & w/o Energy  \\

            \midrule
            
            CLAP$\uparrow$ & 0.44 & 0.236 & 0.237 & 0.52 \\ 
            AV-Align$\uparrow$  & 0.30 & 0.273& 0.272 & 0.10  \\
            
            \bottomrule 
        \end{tabular}
        }
\end{table}

\subsection{Future research directions}
Our architecture is flexible and can also adapt to TTS models, like Diff-TTS~\cite{Jeong2021DiffTTSAD}. Our model could potentially generate nuanced emotional intonation by facial movements, which could have a powerful impact given the limitations in emotional video-to-speech generation. Music generation from dance videos is also promising future research direction.

\subsection{Limitations}
Although our approach successfully leverages the text and video control simultaneously, our method shares the limitation of AudioLDM, namely, hallucination in generated samples. For example, for a given ``basketball bounce'' video, \ours generates a squeaking sound, even if the player is standing still.
We conduct a simple human evaluation on 100 random generated audio samples by \ours on a 5-point scale (1 = no hall., 5 = significant hall.). We got 2.1 average score, with hallucinations identified in 18\% of cases.
This problem might be mitigated if we can use a better AudioLDM model. 


\clearpage
\twocolumn[
\begin{center}
\addcontentsline{toc}{section}{References}
\end{center}
]
\bibliography{aaai25}


\end{document}